\documentclass{article}

\PassOptionsToPackage{numbers}{natbib}
\usepackage[final]{neurips_2019_ml4ad}

\usepackage[utf8]{inputenc} 
\usepackage[T1]{fontenc}    
\usepackage{hyperref}       
\usepackage{url}            
\usepackage{booktabs}       
\usepackage{amsfonts}       
\usepackage{nicefrac}       
\usepackage{microtype}      
\usepackage{blindtext}
\usepackage{color}
\usepackage{siunitx}
\usepackage{multirow}
\usepackage{pgfplots}
\usepackage{amsmath}
\usepackage[linesnumbered,ruled]{algorithm2e}
\usepackage{booktabs}
\usepackage{todonotes}
\DeclareSIUnit\px{px}

\usepackage{pifont}
\newcommand{\cmark}{\ding{51}}%
\newcommand{\xmark}{\ding{55}}%

\title{Visibility Guided NMS: Efficient Boosting of Amodal Object Detection in Crowded Traffic Scenes}

\author{%
Nils Gählert \\ 
Mercedes-Benz AG, R\&\! D \\
University of Jena\\
\texttt{nils.gaehlert@daimler.com} \\
\And
Niklas Hanselmann\\
Mercedes-Benz AG, R\&\! D \\
\texttt{niklas.hanselmann@daimler.com} \\
\And
\AND
Uwe Franke\\
Mercedes-Benz AG, R\&\! D\\
\texttt{uwe.franke@daimler.com} \\
\And
Joachim Denzler\\
University of Jena\\
\texttt{joachim.denzler@uni-jena.de} \\
}
\begin{document}

\maketitle

\begin{abstract}
Object detection is an important task in environment 
perception for autonomous driving. Modern 2D object 
detection frameworks such as Yolo, SSD or Faster R-CNN predict multiple 
bounding boxes per object  that are refined using Non-Maximum Suppression (NMS) to suppress 
all but one bounding box. While object detection itself is fully 
end-to-end learnable and does not require any manual parameter selection, standard NMS 
is parametrized by an overlap threshold that has to be chosen by hand. 
In practice, this often leads 
to an inability of standard NMS strategies to distinguish different 
objects in crowded scenes in the presence of high mutual occlusion, e.g. for parked cars or 
crowds of pedestians.
Our novel Visibility Guided NMS (vg-NMS) leverages both 
pixel-based as well as amodal object detection paradigms and improves the 
detection performance especially for highly occluded objects with little 
computational overhead. We evaluate vg-NMS using KITTI, VIPER as well as the 
Synscapes dataset and show that it outperforms current state-of-the-art NMS. 
\end{abstract}

\section{Introduction}
\begin{figure}[t]
	\centering
	\includegraphics[width=0.49\textwidth]{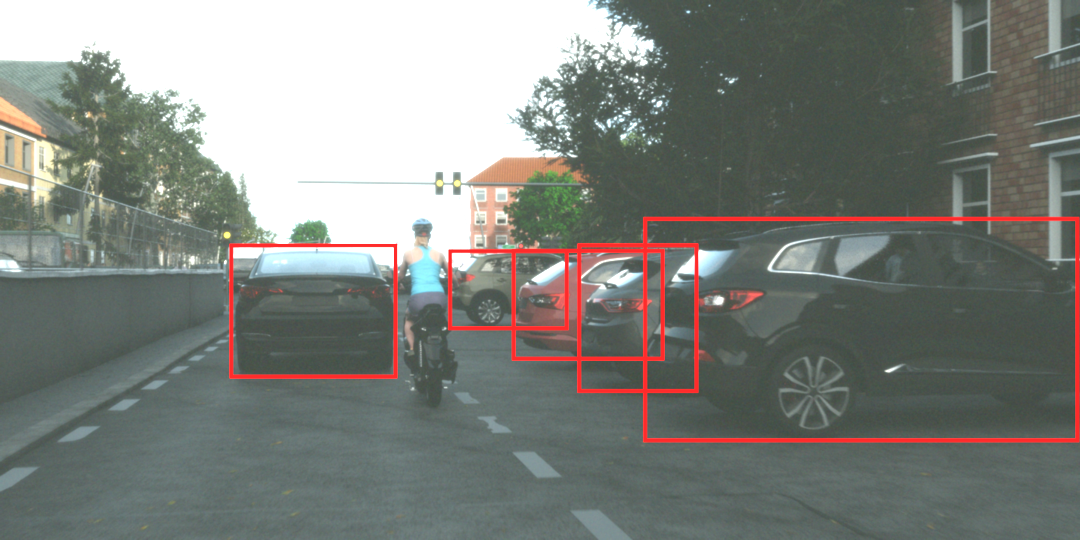}
	\includegraphics[width=0.49\textwidth]{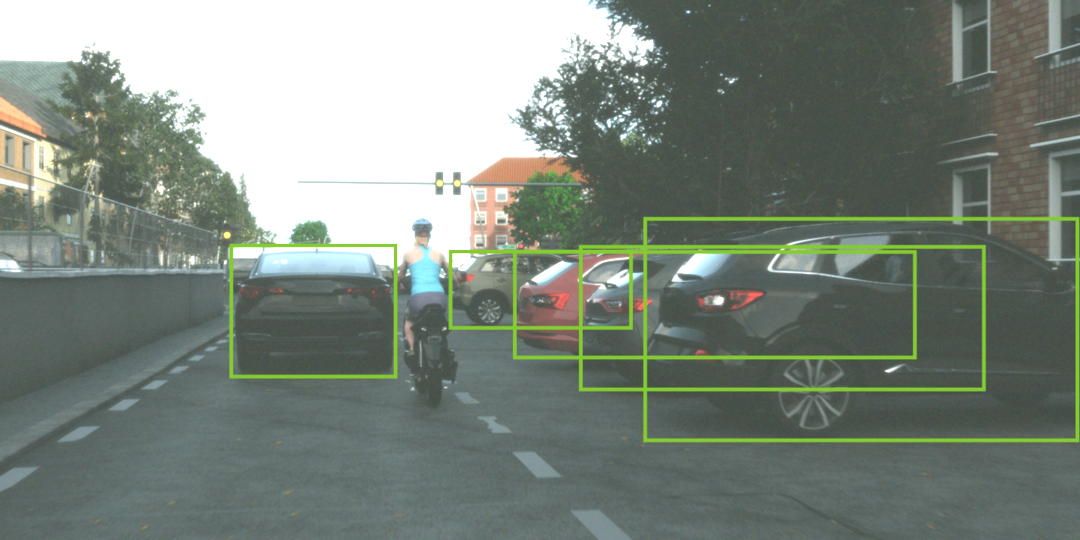}
	\caption{\textbf{Left:} Pixel-based bounding boxes. These bounding boxes 
	only surround visible parts in the image. \textbf{Right:} Amodal bounding 
	boxes. The full extend of the vehicles is surrounded by the bounding box 
	providing a richer representation of the object. Image taken from the 
	Synscapes dataset \cite{wrenninge2018synscapes}\label{fig:tightfull}.}
\end{figure}

Object detection is the task of drawing a tight bounding box around 
an object of interest in an image while still covering all parts of this
object. There are two possible ways to perform object detection: one 
option is to predict the pixel-based box which is restricted to only those 
parts of an object that are visible in the image. The other possibility is to 
predict a box covering the entire object -- even if it is 
not fully visible in the image, for example due to partial occlusion by other 
objects. Humans perceive the environment in the latter way, an ability 
called \emph{amodal perception}, which enables enhanced 
reasoning about a scene as it carries additional information on 
the geometry and full spatial extent of a given object. 
\autoref{fig:tightfull} illustrates the 
difference between these two paradigms.

Independent of which object detection paradigm is chosen, the majority of the 
modern object detectors -- such as Yolo \cite{redmon2016you, 
redmon2017yolo9000, redmon2018yolov3}, SSD \cite{liu2016ssd, fu2017dssd} or 
Faster-RCNN \cite{girshick2015fast, ren2015faster} -- are built upon CNN 
architectures and follow a three-step scheme during inference: (i) a set of 
initial bounding boxes -- known as \emph{prior boxes} or \emph{anchor boxes} -- 
are spread over the whole image. These template boxes are (ii) refined to 
match the outline of the object that is to be detected. Finally, all detections 
are (iii) processed to remove duplicates using 
Non-Maximum Suppression (NMS) \cite{girshick2014rich,girshick2015fast, 
liu2016ssd, redmon2017yolo9000}. While today's object detection frameworks are 
fully end-to-end trainable for steps (i) and (ii), NMS in its commonly used way is still a non-learnable 
algorithm that requires manual parameter tuning. Due to its simplicity, standard NMS is extremly fast while still yielding 
reasonable results for many applications. 

NMS ranks all detections by their score and iteratively removes all candidates
that exceed a manually chosen Intersection-over-Union (IoU) threshold, 
usually between 0.3 and 0.5. 
Detections with a smaller overlap are accepted as 
valid bounding boxes. 
However, as shown in \autoref{fig:standard_nms}, there 
often exist many objects whose respective bounding boxes overlap more strongly 
than this carefully chosen threshold, but which are valid nonetheless 
and would falsely be suppressed. This behaviour can lead to problems 
e.g. in crowded traffic scenarios as by design some objects can never be 
detected, even if the network predicted them correctly.

One solution to this problem is to predict pixel-based bounding boxes, as they 
tend to have lower bounding box overlap for mutually occluded objects than 
their amodal counterparts. However, this induces a loss of information as the 
full spatial extent of the object is no longer considered. 

Our goal is to improve amodal object detection and its relatively 
poor detection performance. We summarize our contributions as follows: (i) We 
propose a novel NMS variant that we call \emph{Visibility Guided NMS} (vg-NMS). 
For vg-NMS only little computational overhead is required as it is built upon 
joint pixel-based and amodal object detection. In contrast to other 
approaches no additional subnetworks are required. (ii) Due to its simple and 
efficient design, vg-NMS can be included in any object detector that features 
standard NMS. (iii) We evaluate vg-NMS on different and challenging datasets 
tailored especially for autonomous driving and show that it outperforms 
standard NMS and its variants.

\section{Related Work}
While amodal perception is a well addressed problem in instance segmentation 
\cite{li2016amodal, zhu2017semantic, follmann2019learning, ehsani2018segan} 
only little attention is paid to amodal object detection in 2D. However, even 
instance segmentation can benefit from amodal object detection e.g. when using 
Mask R-CNN \cite{he2017mask}  as an amodal instance segmentation framework. 

For pure object detection \citet{deng2017amodal} show the potential of 
successfully lifting amodal 2D object detections to 3D space in an indoor 
setting using the NYU v2 dataset \cite{couprie2013indoor} using RGB-D data. 

\begin{figure}[t]
	\centering
	\includegraphics[width=0.49\textwidth]{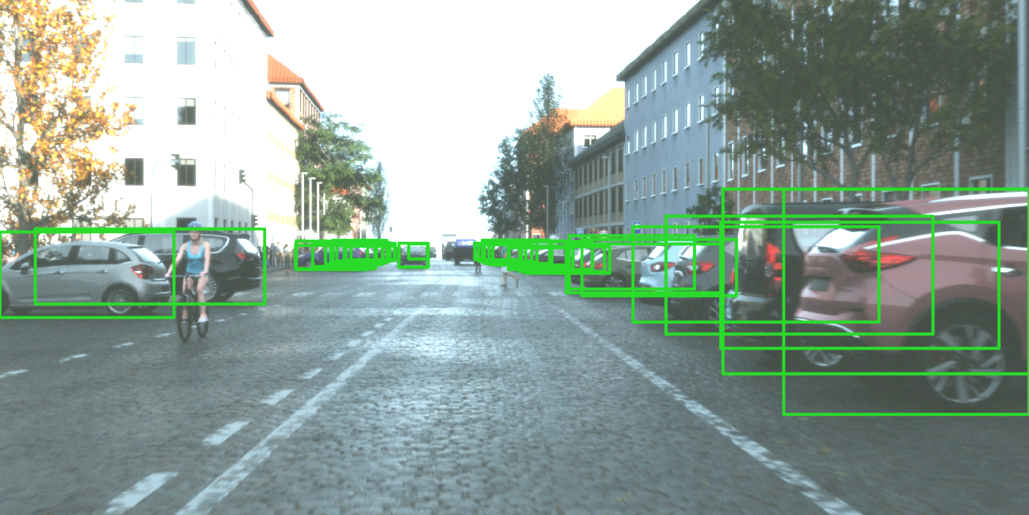}
	\includegraphics[width=0.49\textwidth]{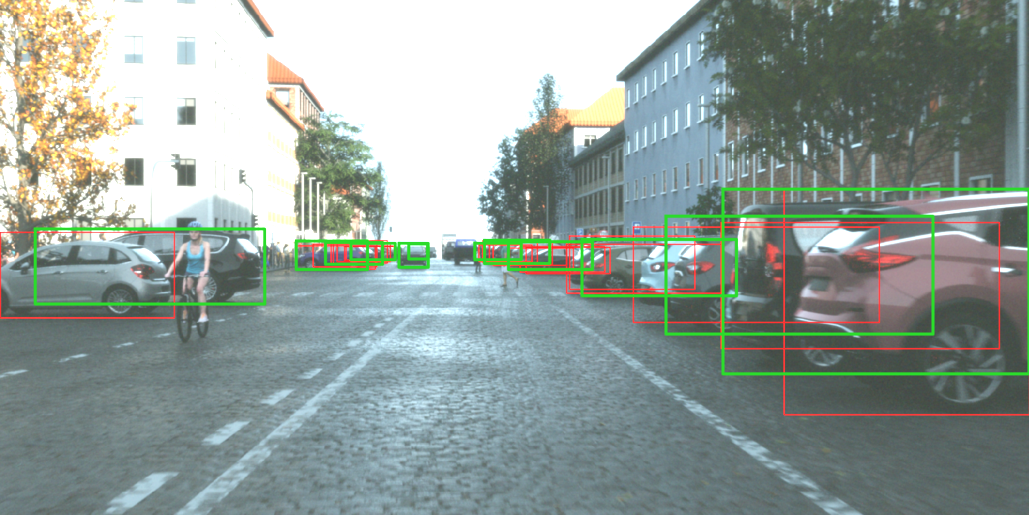}
	\caption{\textbf{Left:} Amodal GT data. \textbf{Right:} If standard NMS is 
	applied on the GT data, several boxes (red) will be removed leading to a 
	bad detection performance. Image taken from the Synscapes dataset 
	\cite{wrenninge2018synscapes}\label{fig:standard_nms}.}
\end{figure}

Especially in an autonomous driving context, several approaches make use of 
amodal object detection. The authors of \cite{mousavian20173d, gahlert2018mb, 
gahlert2019beyond, weberdirect} use monocular RGB images to detect 3D 
vehicles and estimate their orientation. To this end these approaches feature 
an amodal object detection framework and the final 2D detections are enriched 
with additional 3D information. \citet{gahlert2018mb, gahlert2019beyond} build 
on specific keypoints in 2D that -- given the camera projection matrix -- allow
for 3D estimates. All approaches are evaluated on the KITTI Object Detection 
and Orientation Estimation benchmark \cite{geiger2012we}. 
However, most of these methods use 2D object detectors that feature standard 
NMS which tends to perform poorly on amodal bounding boxes compared to 
pixel-based ones as shown in \autoref{fig:standard_nms}. 

There are several approaches addressing NMS related issues. 
Soft NMS \cite{bodla2017soft} ranks candidates by their scores after re-weighting them depending
on their IoU overlap with close-by predictions. 
\citet{he2018softer,he2019bounding} propose a novel bounding box 
regression loss that improves detection performance in combination with a 
variant of Soft NMS called Softer NMS. Fitness NMS \cite{tychsen2018improving} include a 
\emph{fitness} coefficient that measures the IoU overlap between the 
prediction and any GT bounding box.

\citet{liu2019adaptive} improve pedestrian detection performances in a crowd by 
estimating the object density in the image and applying dynamic object 
suppression during NMS. To this end, an additional density estimator subnet is 
required.

Another approach of overcoming issues induced by NMS is to model NMS as a fully 
end-to-end learnable part within the actual object detection framework. 
\citet{hosang2016convnet, hosang2017learning} convert the NMS scheme into an 
additional CNN-subnet, finally making NMS learnable. 

However, appending additional networks like \citet{hosang2017learning} and \citet{liu2019adaptive} results in longer inference time which reduces their merit for 
fully autonomous driving where compute is limited and realtime capable 
algorithms are crucial. Hence, we focus on optimizing NMS with as little 
computational overhead as possible to allow for productive in-vehicle use. 
Contrary to \citet{liu2019adaptive}, we do not only focus on pedestrians and 
extend detection to Cars/Vans and Trucks/Buses.

\section{Joint Pixel-Based and Amodal Object Detection}
To allow for simultaneous amodal and pixel-based object detection several 
straight forward adjustments are required in the network as well as the loss 
design. 

We explain the modifications exemplarily for the Single Shot Multibox Detector 
(SSD) \cite{liu2016ssd} framework, a one-stage detector that performs 
object classification as well as bounding box regression in one step. 

SSD features a CNN backbone network and places multiple prior boxes -- template bounding boxes of different sizes and aspect ratios  -- that are placed regularly all over 
the image.
During training, these prior boxes are trained to match the outlines of 
the object of interest. During inference for each box the class probabilities 
$p_i$ as well as the adjustments relative to the prior box $\delta_x, 
\delta_y, \delta_w, \delta_h$ will be predicted. For each prior box there are 
$4 + (n_c +1)$ free parameters with $n_c$ being the number of different classes. The numbers of prior boxes vary depending on 
the image size. E.g. for KITTI there are approx. \num{11000} prior boxes in our 
setting while VIPER with a resolution of \SI{1920 x 1080}{\px} already has 
nearly \num{50000} prior boxes.
\subsection{Adjustments in Object Detection Framework}

\paragraph{Architecture:}To combine pixel-based and amodal object detection, 
slight adjustments in the network architecture are required. As each prior box 
is now asked to predict two bounding boxes, we add 4 additional regression 
parameters to each prior box leading to $4+4+(n_c +1)$ free parameters per 
prior box. The first set of regression parameters is trained to predict the 
pixel-based bounding boxes while the second set focuses on the amodal bounding 
boxes. 

Compared to the standard SSD setting, these adjustments therefore introduce only a
slight overhead of $n \times 4$ parameters with $n$ being the number of priors per 
image. 

During training, the matching between prior and ground truth boxes is solely 
done using the first set of regression parameters corresponding to the pixel-based 
bounding boxes. 

\paragraph{Loss:} Usually object detection frameworks feature a classification 
and box regression i.e. localization loss
\begin{align}
L_\text{total} = L_\text{cls} + \alpha L_\text{loc}
\end{align}
where $\alpha$ denotes a weighting factor between the two losses. Common 
localization losses are e.g. (Smooth) L1, L2, IoU or gIoU 
\cite{rezatofighi2019generalized} loss. For classification tasks  
Cross-Entropy Loss (CE) or its improved version Focal Loss \cite{lin2017focal} 
are commonly used.

For joint detection two different objectives must be optimized for -- i.e. 
the regression of pixel-based and amodal bounding boxes. The overall 
localization loss is thus composed of two terms:
\begin{align}
L_\text{loc} &= \beta L_\text{loc,pix} + \gamma L_\text{loc,amodal}. \\
\intertext{Hence the total loss is given by}
L_\text{total} &= L_\text{cls} + \beta  L_\text{loc,pix} + \gamma L_\text{loc,amodal}.
\end{align}
In our experiments we use $\alpha = 1$ for networks that only perform 
one detection task. For joint pixel-based and amodal object detection we find  
$\beta  =1 $ and $\gamma = 2$ to be a good combination.

\subsection{Visibility Guided NMS}
To ensure that both pixel-based as well as 
amodal prior boxes at index $i$ will cover the same physical object, we employ 
element-wise sharing of the classifier weights between both sets of prior boxes
$\mathcal P_\text{pix}$ and $\mathcal P_\text{amodal}$.  An object covered by the corresponding prior boxes
will therefore yield the same classification result. 

The main idea of Visibility Guided NMS (vg-NMS) is to perform NMS on the pixel-based 
bounding boxes that describe the actually visible parts of the object, but 
output the amodal bounding boxes that belong to the indices that are retained 
during pixel-based NMS. For a set of initial detections $\mathcal B$, standard 
NMS returns a set of indices $\mathcal I$ containing all indices of detections 
in $\mathcal B$ that are retained during the NMS. Finally the corresponding 
boxes in $\mathcal B$ are selected by indices $\mathcal I$ yielding the final set of detections $\mathcal D$.

In contrast, vg-NMS combines standard NMS with the two different sets of object predictions 
$\mathcal B_\text{pix}$ and $\mathcal B_\text{amodal}$ and performs standard NMS on $\mathcal B_\text{pix}$ returning a set of indices $\mathcal 
I_\text{pix}$. The final valid detections both for pixel-based $\mathcal D_\text{pix}$ and amodal bounding boxes $\mathcal 
D_\text{amodal}$ are chosen by $\mathcal I_\text{pix}$. 

\input{nms_illustration}

\autoref{fig:visu_hybridnms} and \autoref{algo:hybrid} illustrate vg-NMS both 
visually and as pseudocode.

\begin{algorithm}[t]
    \SetKwInOut{Input}{Input}
    \SetKwInOut{Output}{Output}

    \underline{VisibilityGuidedNMS} $(\mathcal B_\text{pix}, \mathcal B_\text{amodal})$\;
    \Input{Set of predictions $\mathcal B_\text{pix}$, $\mathcal B_\text{amodal}$}
    \Output{Set of detections $\mathcal D_\text{pix}$, $\mathcal D_\text{amodal}$}
    
    $\mathcal I_\text{pix} \leftarrow$ NMS$(\mathcal B_\text{pix})$\;
    $\mathcal D_\text{pix} \leftarrow$ select$(\mathcal B_\text{pix}, \mathcal I_\text{pix})$\;
    $\mathcal D_\text{amodal} \leftarrow$ select$(\mathcal B_\text{amodal}, \textcolor{red}{\mathcal I_\text{pix}})$ \tcc*[r]{$\mathcal I_\text{pix} $ is used here}
    \Return{$\mathcal D_\text{pix}$, $\mathcal D_\text{amodal}$} 
    \caption{vg-NMS. \label{algo:hybrid}}
\end{algorithm}

\section{Experiments}
\subsection{Datasets and Evaluation Metrics}
Several datasets exist that are tailored specifically for autonomous driving 
purposes. However, most of the datasets focus on specific tasks such as 
semantic and instance segmentation or object detection. An overview of 
currently available datasets is shown in \autoref{tab:dataset}. Only few 
datasets feature amodal object detection and even less datasets feature both 
amodal and pixel-based object detection. Fortunately, if instance segmentation 
data is available, pixel-based bounding boxes can easily be generated. 

For our experiments we select KITTI \cite{geiger2012we}, Viper 
\cite{richter2017playing} and Synscapes \cite{wrenninge2018synscapes} as these 
datasets contain annotations for both amodal as well as pixel-based bounding 
boxes. KITTI contains nearly \num{15000} real images with a resolution of 
approx. \SI{380 x1250}{\px}. Synscapes (\SI{720 x1440}{\px}) and VIPER 
(\SI{1080 x 1920}{\px}) are synthetic datasets with outstanding and nearly 
photorealistic graphics. Synscapes contains \num{25000} images and VIPER, the 
most exhaustive resource of the three, offers more than \num{250000} images. 

For KITTI we follow \cite{xiang2015data} to create a train-val-split. This
ensures that no images from the same sequence are seen during both 
training and validation. Please note that KITTI does not 
provide official semantic and instance segmentation data. Instead we use KINS 
\cite{qi2019amodal}, which extends KITTI with the corresponding 
labels. Unfortunately, the label definitions used for KINS are not in full concordance with the  
 definitions used in the original KITTI dataset.
 
VIPER offers an official train-val-split. For Synscapes, we selected the first 
\SI{80}{\percent} of the \num{25000} images as a training set and hold out the remaining 
\SI{20}{\percent} for validation. All experiments were evaluated on 
the respective validation sets. 

As all datasets follow different labeling strategies we select three shared 
classes with matching definitions accross all datasets: (i) cars \& vans, (ii) 
trucks \& buses and (iii) pedestrians. Motorbikes and bicycles are not labeled 
in all datasets, hence we do not consider these objects to allow for a valid 
comparison.

While for Synscapes and VIPER there is no official object detection benchmark available, KITTI provides such a benchmark. 
However, as the label definitions as used within our 
experiments do not match the exact definitions for this benchmark, we did not evaluate 
on the official KITTI benchmark.
\begin{table}[t]
  \caption{Current datasets tailed especially for autonomous driving. For object detection not all datasets feature both pixel-based as well amodal bounding boxes.}
  \label{tab:dataset}
  \centering
\begin{tabular}{lrccc}
  \toprule
  &  & \multicolumn{3}{c}{Annotations}          \\ \cmidrule{3-5}
Name& Real & Pixel-based 2D Box & Amodal 2D Box & 3D Box \\ \midrule
KITTI \cite{geiger2012we} & \cmark & \cmark\ (KINS \cite{qi2019amodal})&\cmark&\cmark\\
Cityscapes \cite{cordts2016cityscapes} & \cmark &\cmark& \xmark &\xmark\\
BDD100k \cite{yu2018bdd100k}&\cmark&\cmark&\xmark&\xmark\\
ApolloScape \cite{huang2018apolloscape}&\cmark&\cmark&\xmark&\cmark\ (Lidar)\\
Mapillary \cite{neuhold2017mapillary}&\cmark&\cmark&\xmark&\xmark\\
Nuscenes \cite{nuscenes2019}&\cmark&\xmark&\cmark&\cmark\\
Lyft \cite{lyft2019}&\cmark&\xmark&\cmark&\cmark\\
Waymo Open Dataset \cite{waymo_open_dataset}&\cmark&\xmark & \cmark&\cmark\\
\midrule
Synthia \cite{ros2016synthia} &\xmark&\cmark&\cmark&\cmark\\
VIPER \cite{richter2017playing}&\xmark&\cmark&\cmark&\cmark\\
Synscapes \cite{wrenninge2018synscapes} &\xmark&\cmark&\cmark&\cmark\\

\bottomrule
\end{tabular}
\end{table}

As shown in \autoref{fig:standard_nms} in a standard NMS setting, all objects 
with an overlap greater than a specific threshold -- e.g. $0.45$ -- cannot be resolved. However, especially for 
vehicles there is a higher fraction of objects with high overlap in amodal 
object bounding boxes compared to pixel-based bounding boxes. By using vg-NMS 
the non-resolvable objects will decrease to the pixel-based fraction and thus 
will yield better results.

For each dataset we train a network specialized for the task of detecting 
amodal bounding boxes and compare these with a network that features 
joint pixel-based and amodal object detection and vg-NMS.

We use Mean Average Precision (mAP) as the evaluation metric and exclude all 
boxes with a size of smaller than \SI{20 x20}{\px} as detecting small objects 
is a challenging task for all object detectors \cite{cao2018feature}. Other benchmarks use similar exclusion criteria, e.g. KITTI excludes all boxes with a size of smaller than \SI{25 x25}{\px}. In Cityscapes \cite{cordts2016cityscapes} all objects with a size of less than  \SI{10 x10}{\px} are not taken into account during evaluation.
Additionally, IoU calculation is very unstable for very small objects leading 
to a higher number of False Positives (FP) as well as False Negatives (FN).

Other evaluation methods like KITTI \cite{geiger2012we} also ignore heavily 
occluded or truncated boxes. As vg-NMS specifically addresses this scenario, we 
do not exclude these boxes to allow for a realistic 
assessment of its performance. Furthermore, we require an IoU overlap of $0.5$ 
to accept a detection as a True Positive (TP).

\subsection{Theoretical Problem Analysis}
Each dataset is evaluated to quantify the theoretically possible improvement 
for detection performance for both pedestrians as well as vehicles including 
cars, vans, trucks and buses. 

As standard NMS removes objects instead of spawning new ones, the number of 
False Negatives (FN) will increase yielding a decreased Recall value $R = \tfrac {TP}{TP + FN}$. The 
maximal theoretically possible Recall values are shown in 
\autoref{tab:theo_iou} and an in-depth analysis is given in the appendix 
\ref{app:theo}. 


\begin{table}
  \caption{Theoretically improvements of vg-NMS. The maximum possible Recall values $R_\text{max}$ and $R_\text{vg}$ are given for the standard NMS as well as the vg-NMS. Furthermore the number of objects per image is given. We only consider boxes with a size of at least \SI{20 x 20}{\px} as smaller objects are hard to identify even for humans. Especially for vehicles there is huge potential of theoretical improvments by using vg-NMS. }
  \label{tab:theo_iou}
  \centering
\begin{tabular}{lrrrrrrrr}
\toprule
          & \multicolumn{4}{c}{Vehicles} & \multicolumn{4}{c}{Pedestrians} \\ \cmidrule(l{0pt}r{2pt}){2-5}\cmidrule(l{2pt}r{0pt}){6-9}
          & $R_\text{max}$  & $R_\text{vg}$& $\Delta$ [\%]  &\#/image & $R_\text{max}$ & $R_\text{vg}$& $\Delta$ [\%]  &\#/image\\ \midrule
KITTI \cite{geiger2012we}& \num{0.956}&\num{0.965}&\textbf{+\num{0.94}}&\num{2.9}&\num{0.978}&\num{0.982}&\textbf{+\num{0.41}}&\num{0.4}\\
VIPER \cite{richter2017playing}&\num{0.948}&\num{0.981}&\textbf{+\num{3.48}}&\num{3.7}&\num{0.951}&\num{0.980}&\textbf{+\num{3.05}}&{1.4}\\
Synscapes \cite{wrenninge2018synscapes}&\num{0.869}&\num{0.939}&\textbf{+\num{8.05}}&\num{7.1}&\num{0.918}&\num{0.952}&\textbf{+\num{3.75}}&\num{7.9}\\
\bottomrule
\end{tabular}
\end{table}

While KITTI is the only dataset used in this study that features real-world 
data, it is the dataset with the smallest possible improvements compared to 
VIPER and Synscapes. This is because KITTI includes only a few valid 
objects per image compared to VIPER and Synscapes since it includes 
traffic scenes from both urban and overland environments. The latter scenario
of course features crowded scenes much less frequently.

A more reasonable real-world dataset for the evaluation on crowded traffic 
scenes is Cityscapes \cite{cordts2016cityscapes} as it mainly focuses on urban 
environments. However, with an average of \num{4.2} and \num{8.4} bounding 
boxes per image for the pedestrian and vehicle class respectively, the typical scene 
composition in Cityscapes is similar to the Synscapes dataset, making it an 
adequate surrogate for our purposes.

Please note that these theoretical values can only be reached if the underlying 
object detector perfectly predicts all bounding boxes. In practice 
these numbers may therefore vary or $R_\text{max}$ as well as 
$R_\text{vg}$ might not be reached at all.

\subsection{Results}

\begin{table}
  \caption{Average Precision (AP) results of different NMS variants for amodal object detection. Evaluation was performed on the validation splits of each dataset. vg-NMS outperforms both standard and Soft NMS and shows excellent performance in combination with Soft NMS.}
  \label{tab:map_results_gt20px}
{\centering
\begin{tabular}{llrrrrr}
\toprule
                       &       & \multicolumn{4}{c}{AP [\%]} & \\  \cmidrule{3-6}
                       &                   & Car / Van & Truck / Bus & Pedestrian & mAP & Runtime* \\ \midrule
\multirow{4}{*}{KITTI \cite{geiger2012we}} & Standard NMS      &  \num{80.26}&\textbf{\num{37.12}}&\num{62.57}&\num{59.98}  &\textbf{\SI{21.2}{\milli\second}}\\ \cmidrule{2-7} 
                       & Soft         & \num{79.86}&\num{36.88}&\num{62.24}&\num{59.66}  &\SI{71.3}{\milli\second}\\ 
                       & vg        & \textbf{\num{80.65}}&\num{36.46}&\textbf{\num{63.74}}&\textbf{\num{60.29}}  &\SI{22.8}{\milli\second}\\ 
                       & vg + Soft  &\num{80.30}&\num{36.54}&\num{63.51}&\num{60.12} &\SI{75.2}{\milli\second}\\
\midrule
\multirow{4}{*}{VIPER \cite{richter2017playing}} & Standard NMS      &\num{72.28}&\num{33.30}&\num{51.49}&\num{52.36} &\textbf{\SI{35.5}{\milli\second}}\\ \cmidrule{2-7} 
                       & Soft           & \num{72.40}&\num{33.34}&\num{51.46}&\num{52.40} &\SI{85.3}{\milli\second}\\
                       & vg         & \num{72.98}&\num{33.47}&\num{54.78}&\num{53.74}&\SI{37.4}{\milli\second}\\ 
                       & vg + Soft  & \textbf{\num{73.04}}&\textbf{\num{33.53}}&\textbf{\num{54.80}}&\textbf{\num{53.79}} &\SI{68.4}{\milli\second}\\
\midrule
\multirow{4}{*}{Synscapes \cite{wrenninge2018synscapes}} & Standard NMS     & \num{83.97}&\num{84.62}&\num{87.33}&\num{85.31}&\textbf{\SI{26.7}{\milli\second}}\\ \cmidrule{2-7} 
                       & Soft         & \num{83.66}&\num{84.59}&\num{86.99}&\num{85.08}&\SI{113.1}{\milli\second}\\
                       & vg          & \num{88.25}&\num{85.61}&\textbf{\num{89.68}}&\textbf{\num{87.85}}&\SI{28.2}{\milli\second}\\ 
                       & vg + Soft  &  \textbf{\num{88.26}}& \textbf{\num{85.63}}&\num{89.60}&\num{87.83}&\SI{78.8}{\milli\second}\\
\bottomrule
\end{tabular}}
\footnotesize{\newline \newline  * For Soft NMS we use TensorFlow's built-in function \verb|non_max_suppression_v5| which is still under development and not yet publically available in an official release. We suspect that due to the implementation runtime of Soft NMS increases with the number of box predictions with high IoU overlap. In case of amodal object detection much more predictions have a high overlap, hence Soft NMS is much slower for amodal bounding boxes compared to pixel-based ones. E.g. for a subset of 100 Synscapes images we find in average \num{2.35} times more amodal bounding boxes with an IoU overlap of more than \num{0.45} compared to pixel-based bounding boxes.}
\end{table}
All experiments were conducted under the same setup. We use TensorFlow and 
InceptionV1 \cite{szegedy2015going} as a backbone for our SSD network. We 
trained the networks using the Adam optimizer \cite{kingma2014adam} with a 
learning rate of $lr=0.0001$, $\beta_1 = 0.9$, $\beta_2=0.999$ and a batch size of 1. Every training was performed multiple times with different random seeds and evaluated after each epoch using the validation set. 
Finally, the results of the best snapshots are reported.
During training we choose L2 as the bounding box 
regression loss $L_\text{loc}$ for both pixel-based and amodal bounding boxes. 
Furthermore, we use Focal Loss \cite{lin2017focal} as the classification task 
loss. Both training and inference were performed on an NVIDIA Titan RTX. 

\autoref{tab:map_results_gt20px} shows the results for the standard NMS setting 
as well as for Soft NMS and vg-NMS. Qualitative results of vg-NMS are shown in 
the appendix \ref{sub:vis_results}. 

For all datasets vg-NMS outperforms standard NMS and Soft NMS. The combination 
of Soft and vg-NMS often yields even better results. Especially for VIPER and 
Synscapes the detection performance was significantly increased. This is in 
concordance to \autoref{tab:theo_iou} where we expect the most gain in 
performance primarily for datasets with crowded scenes -- i.e. many overlapping 
objects in one image. 

While detection performance is improved, the runtime of vg-NMS remains comparable or even decreases: On 
average, vg-NMS is just \SI{6.2}{\percent} slower when paired with a standard hard suppresion 
strategy. This amounts to less than \SI{2}{\milli\second} computational overhead due to the simultanous estimation 
of both pixel-based and amodal bounding boxes. Furthermore, vg-NMS is 
capable of significantly reducing the overall runtime if used in combination 
with Soft NMS for both VIPER and Synscapes images. We suspect that this is 
because pixel-based boxes inherently overlap less strongly with each other than 
their amodal counterparts. Hence the number of candidate 
pairings that need to be considered during the score reweighting step in 
Soft NMS is significantly smaller for pixel-based bounding boxes.

\subsection{Ablation Study}
By simultaneously predicting both pixel-based and amodal bounding boxes we de 
facto created a multitask setting as two tasks are optimized in parallel. 
\citet{caruana1997multitask} shows that multitask learning inherently improves 
the overall detection performance as it allows for better generalizations of 
the features that are learned.

Hence, we evaluate all 3 datasets regarding their performance in the single 
task networks and in the multitask setting. For these experiments we use the 
same data as for the previous ones. The results are shown in 
\autoref{tab:multitask}.

As expected, the simultaneous regression of both pixel-based and amodal object 
detection leads to an improved single task performance.


\begin{table}
  \caption{Single task Average Precision (AP) results for standard NMS for both single task and multitask networks. Evaluation was performed on the validation splits of each dataset. Regressing both pixel-based and amodal bounding boxes jointly leads to an improvement compared the single task setting.}
  \label{tab:multitask}
  \centering
\begin{tabular}{lllrrrr}
\toprule
                       &   &    & \multicolumn{4}{c}{AP [\%]} \\  \cmidrule{4-7}
                       & Box type & Network                  & Car / Van & Truck / Bus & Pedestrian & mAP  \\ \midrule
\multirow{4}{*}{KITTI \cite{geiger2012we}} & Pixel      & Single& \num{78.37}&\textbf{\num{36.89}}&\textbf{\num{63.18}}&\textbf{\num{59.51}} \\
                       & Pixel      & Multi& \textbf{\num{79.24}}&\num{36.25}&\num{62.71}&\num{59.40} \\  \cmidrule{2-7} 
                       & Amodal     & Single& \textbf{\num{80.26}}&\textbf{\num{37.12}}&\num{62.57}&\num{59.98} \\  
                       & Amodal    & Multi& \textbf{\num{80.26}}&\num{36.58}&\textbf{\num{63.54}}&\textbf{\num{60.12}} \\ 
\midrule
\multirow{4}{*}{VIPER \cite{richter2017playing}} & Pixel      & Single&\num{75.28}&\num{38.41}&\num{48.47}&\num{54.05}\\ 
                       & Pixel      & Multi&\textbf{\num{76.88}}&\textbf{\num{40.48}}&\textbf{\num{49.45}}&\textbf{\num{55.60}}\\  \cmidrule{2-7}
                       & Amodal     & Single&\num{72.28}&\textbf{\num{33.30}}&\num{51.49}&\num{52.36}\\ 
                       & Amodal      & Multi&\textbf{\num{72.39}}&\num{33.25}&\textbf{\num{55.32}}&\textbf{\num{53.65}}\\ 
 \midrule

\multirow{4}{*}{Synscapes \cite{wrenninge2018synscapes}}  & Pixel      & Single& \num{84.25}&\num{82.72}&\num{86.54}&\num{84.50}  \\ 
                       & Pixel      & Multi& \textbf{\num{85.11}}&\textbf{\num{83.58}}&\textbf{\num{87.26}} & \textbf{\num{85.32}}  \\ \cmidrule{2-7} 
                       & Amodal     & Single& \num{83.97}&\num{84.62}&\num{87.33}&\num{85.31}  \\ 
                       & Amodal      & Multi& \textbf{\num{85.41}}&\textbf{\num{85.84}}&\textbf{\num{88.76}}&\textbf{\num{86.67}}  \\  
\bottomrule
\end{tabular}
\end{table}

\section{Conclusion}
In this paper a new variant of Non-Maximum Suppression (NMS) called \emph 
{Visibility Guided NMS} (vg-NMS) was proposed. vg-NMS can be used to boost 
amodal object detection in crowded traffic scenes as it prevents highly 
occluded amodal bounding boxes from being falsely suppressed. In contrast to 
other techniques we do not require an additional subnet but only need labeled 
GT data for both pixel-based as well as amodal bounding boxes. While vg-NMS 
itself does not need any additional computation, slight adjustments in the 
network design are required to allow for a simultaneous detection of both types 
of bounding boxes. In average these adjustments cause an increase of runtime of 
only \SI{6.2}{\percent}. Furthermore, we showed that simultaneous pixel-based 
and amodal bounding object detection also leads to an improved performance in 
the single task setting without vg-NMS.

vg-NMS can be included in all modern object detectors such as SSD 
\cite{liu2016ssd}, YOLO \cite{redmon2017yolo9000} or Faster RCNN 
\cite{ren2015faster}. We evaluated vg-NMS on KITTI, VIPER and Synscapes -- 
datasets tailored especially for autonomous driving -- and show that vg-NMS 
outperforms standard NMS as well as Soft NMS especially in crowded scenarios 
with a high number of objects per image. 


\clearpage
\bibliographystyle{plainnat}
\bibliography{bib}

\clearpage
\appendix
\section{Appendix}
\subsection{In-Depth Theoretical Analysis}
\label{app:theo}
For each dataset we analyze the distribution of overlapping objects for all 
vehicles (including cars, trucks and buses) as well as for pedestrians. As 
bicycles and motorbikes are not labeled in each dataset, these vehicles are 
excluded in our analysis. For each image in the dataset we opt for the maximum 
IoU of one bounding box with any other object.

\begin{figure}[t]
 \begin{tikzpicture}
        \begin{axis}[
          ybar interval,
	xmin=0.0,xmax=1,
	ymin=0,ymax=1.5,
	xtick={0,0.2,...,1},
	xticklabels=\empty,	
	grid=both,
	height=5cm,
	width=0.53\textwidth,
	ylabel={$p_\text{KITTI}$ [\%]},
	title={Vehicles (cars, vans, trucks, buses)},
	tick align=inside
        ]
	\addplot [draw=none, fill={rgb, 255:red, 65; green, 117; blue, 5 }] table [x=x, y=y ,col sep = space] {data/ious_kins_vehicle_full.dat};
	\addlegendentry{Amodal}
	\addplot [draw=none, fill={rgb, 255:red, 255; green, 0; blue, 0 }] table [x=x, y=y ,col sep = space] {data/ious_kins_vehicle_tight.dat};
	\addlegendentry{Pixel-based}
	\draw[line width=2,blue] (45,0) -- (45,150);

        \end{axis}
    \end{tikzpicture}
    ~%
    \begin{tikzpicture}
        \begin{axis}[
          ybar interval,
	xmin=0.0,xmax=1,
	ymin=0,ymax=1.5,
	xtick={0,0.2,...,1},
	xticklabels=\empty,	
	yticklabels=\empty,	
	grid=both,
	height=5cm,
	width=0.53\textwidth,
	title={Pedestrians},
	tick align=inside
        ]
	\addplot [draw=none, fill={rgb, 255:red, 65; green, 117; blue, 5 }] table [x=x, y=y ,col sep = space] {data/ious_kins_ped_full.dat};
	\addlegendentry{Amodal}
	\addplot [draw=none, fill={rgb, 255:red, 255; green, 0; blue, 0 }] table [x=x, y=y ,col sep = space] {data/ious_kins_ped_tight.dat};
	\addlegendentry{Pixel-based}
	\draw[line width=2,blue] (45,0) -- (45,150);

        \end{axis}
 \end{tikzpicture}
 \begin{tikzpicture}
        \begin{axis}[
          ybar interval,
	xmin=0.0,xmax=1,
	ymin=0,ymax=1.5,
	xtick={0,0.2,...,1},
	xticklabels=\empty,	
	grid=both,
	height=5cm,
	width=0.53\textwidth,
	ylabel={$p_\text{VIPER}$ [\%]},
	tick align=inside
        ]
	\addplot [draw=none, fill={rgb, 255:red, 65; green, 117; blue, 5 }] table [x=x, y=y ,col sep = space] {data/ious_gta_vehicle_full.dat};
	\addlegendentry{Amodal}
	\addplot [draw=none, fill={rgb, 255:red, 255; green, 0; blue, 0 }] table [x=x, y=y ,col sep = space] {data/ious_gta_vehicle_tight.dat};
	\addlegendentry{Pixel-based}
	\draw[line width=2,blue] (45,0) -- (45,150);

        \end{axis}
    \end{tikzpicture}
    ~\hspace*{4mm}%
    \begin{tikzpicture}
        \begin{axis}[
          ybar interval,
	xmin=0.0,xmax=1,
	ymin=0,ymax=1.5,
	xtick={0,0.2,...,1},
	yticklabels=\empty,	
	xticklabels=\empty,	
	grid=both,
	height=5cm,
	width=0.53\textwidth,
	tick align=inside
        ]
	\addplot [draw=none, fill={rgb, 255:red, 65; green, 117; blue, 5 }] table [x=x, y=y ,col sep = space] {data/ious_kins_ped_full.dat};
	\addlegendentry{Amodal}
	\addplot [draw=none, fill={rgb, 255:red, 255; green, 0; blue, 0 }] table [x=x, y=y ,col sep = space] {data/ious_kins_ped_tight.dat};
	\addlegendentry{Pixel-based}
	\draw[line width=2,blue] (45,0) -- (45,150);

        \end{axis}
    \end{tikzpicture}
 
 \begin{tikzpicture}
        \begin{axis}[
          ybar interval,
	xmin=0.0,xmax=1,
	ymin=0,ymax=1.5,
	xtick={0,0.2,...,1},
	xticklabel style = {xshift=-0.55cm},
	grid=both,
	height=5cm,
	width=0.53\textwidth,
	xlabel={IoU},
	ylabel={$p_\text{Synscapes}$ [\%]},
	tick align=inside
        ]
	\addplot [draw=none, fill={rgb, 255:red, 65; green, 117; blue, 5 }] table [x=x, y=y ,col sep = space] {data/ious_synscapes_vehicle_full.dat};
	\addlegendentry{Amodal}
	\addplot [draw=none, fill={rgb, 255:red, 255; green, 0; blue, 0 }] table [x=x, y=y ,col sep = space] {data/ious_synscapes_vehicle_tight.dat};
	\addlegendentry{Pixel-based}
	\draw[line width=2,blue] (45,0) -- (45,150);

        \end{axis}
    \end{tikzpicture}
    ~\hspace*{4mm}%
    \begin{tikzpicture}
        \begin{axis}[
          ybar interval,
	xmin=0.0,xmax=1,
	ymin=0,ymax=1.5,
	xtick={0,0.2,...,1},
	xticklabel style = {xshift=-0.55cm},
	yticklabels=\empty,	
	grid=both,
	height=5cm,
	width=0.53\textwidth,
	xlabel={IoU},
	tick align=inside
        ]
	\addplot [draw=none, fill={rgb, 255:red, 65; green, 117; blue, 5 }] table [x=x, y=y ,col sep = space] {data/ious_synscapes_ped_full.dat};
	\addlegendentry{Amodal}
	\addplot [draw=none, fill={rgb, 255:red, 255; green, 0; blue, 0 }] table [x=x, y=y ,col sep = space] {data/ious_synscapes_ped_tight.dat};
	\addlegendentry{Pixel-based}
	\draw[line width=2,blue] (45,0) -- (45,150);

        \end{axis}
    \end{tikzpicture}
    \caption{Histograms for vehicles and pedestrians in the VIPER, KITTI and Synscapes dataset for both amodal and pixel-based bounding boxes. Due to the standard NMS design all detections with an IoU of greater than e.g. $0.45$ cannot be resolved. For amodal bounding boxes a higher number of objects will be lost due to NMS.\label{fig:hist_viper}}
\end{figure}
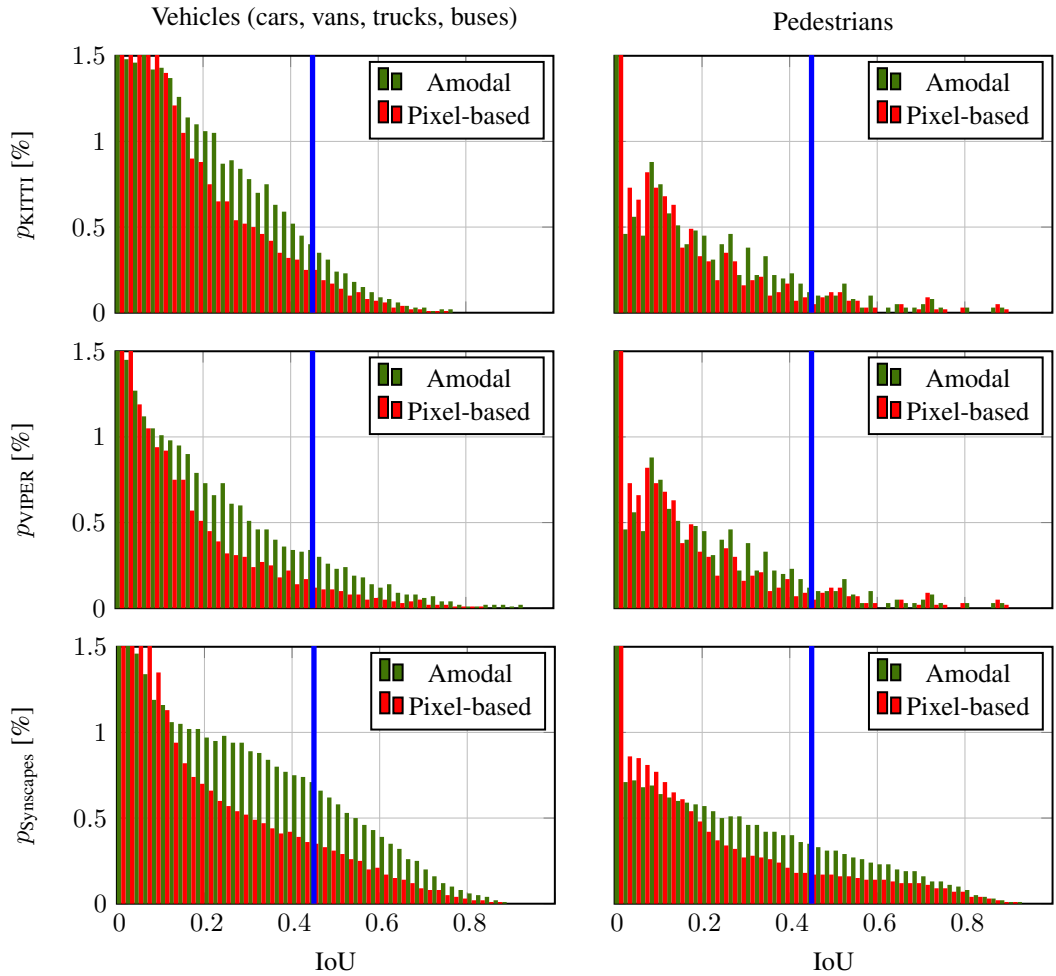

The corresponding histograms are shown in \autoref{fig:hist_viper} and zoomed 
to all IoU overlaps $> 0.05$. All objects above the IoU threshold -- in our 
experiments we set it to $0.45$ -- cannot be resolved. Hence the number of False 
Negatives (FN) will increase. Given the appearance histogram with bins $i \in 
\mathcal B$ with density $p(i)$, IoU threshold $t_\text{IoU}$ we will find for 
the theoretical maximum True Positive (TP), False Negative (FN) and Recall 
($R$) values for amodal object detection
\begin{align}
TP_\text{max} &= 1 - \sum_{i > t_\text{IoU}}^1 p_\text{amodal}(i) \\
FN_\text{max} &= \sum_{i > t_\text{IoU}}^1 p_\text{amodal}(i)\\
R_\text{max} &= \frac {TP_\text{max}}{TP_\text{max} + FN_\text{max}} 
\label{eq:rmax} \\
&= 1 - \sum_{i > t_\text{IoU}}^1 p_\text{amodal}(i). \\
\intertext{For vg-NMS these theoretical upper bounds can be calculated as}
TP_\text{vg} &= 1 - \sum_{i > t_\text{IoU}}^1 
\underbrace{\left(p_\text{amodal}(i) - p_\text{pix}(i)\right)}_{\kappa_i} \\
FN_\text{vg} &= \sum_{i > t_\text{IoU}}^1 \underbrace{\left(p_\text{amodal}(i) - p_\text{pix}(i)\right)}_{\kappa_i} \\
R_\text{vg} &= \frac {TP_\text{vg}}{TP_\text{vg} + FN_\text{vg}} \\
&= 1 - \sum_{i > t_\text{IoU}}^1 \left(p_\text{amodal}(i) - 
p_\text{pix}(i)\right) \\
&= 1 - \sum_{i > t_\text{IoU}}^1 p_\text{amodal}(i) + \sum_{i > t_\text{IoU}}^1 
p_\text{pix}(i) \\
R_\text{vg} &= R_\text{max} + \sum_{i > t_\text{IoU}}^1 p_\text{pix}(i). 
\label{eq:rhybrid}
\end{align}
As shown in \autoref{fig:hist_viper} the coefficient $\kappa_i$ is very small for 
pedestrians compared to vehicles. Hence, vehicles benefit the most of vg-NMS.
An overview about the theoretical improvements is shown in 
\autoref{tab:theo_iou}.


\clearpage
\subsection{Visual Examples}
The following images are taken from the \emph{val} split of each dataset. For KITTI the images are part of the official \emph{test} set.
\label{sub:vis_results}
\subsubsection{KITTI}
\begin{figure}[h]
	\centering
	\includegraphics[width=0.49\textwidth]{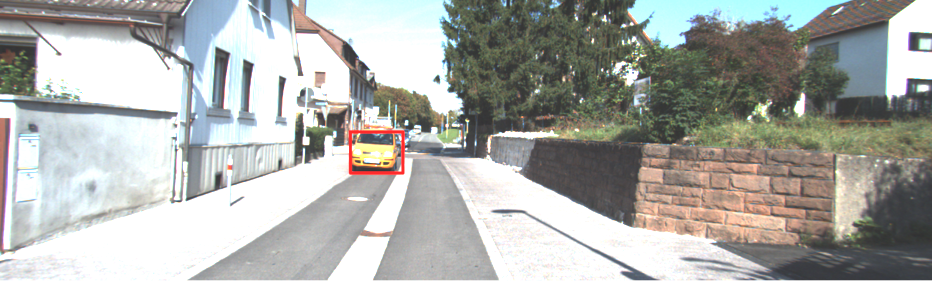}
	\includegraphics[width=0.49\textwidth]{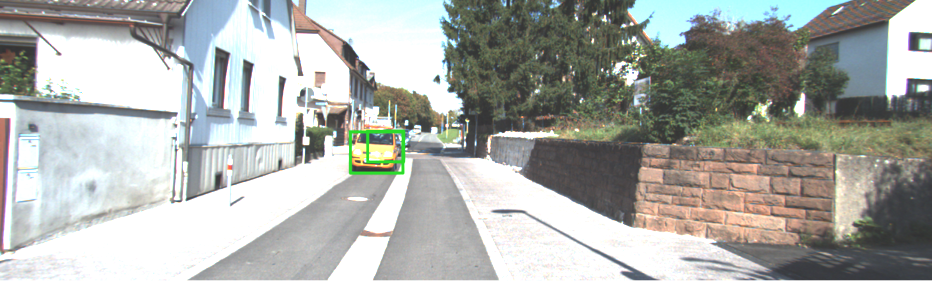}
	\includegraphics[width=0.49\textwidth]{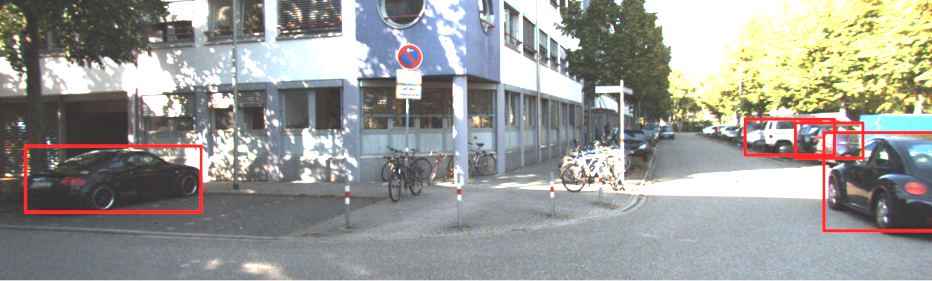}
	\includegraphics[width=0.49\textwidth]{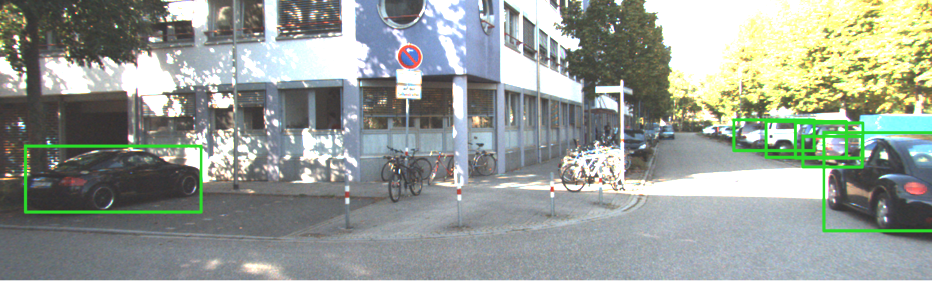}
	\caption{\textbf{Left:} Final detections without vg-NMS. \textbf{Right:} Final detections with vg-NMS. \label{fig:example_result}}
\end{figure}

\subsubsection{VIPER}

\begin{figure}[h]
	\centering
	\includegraphics[width=0.49\textwidth]{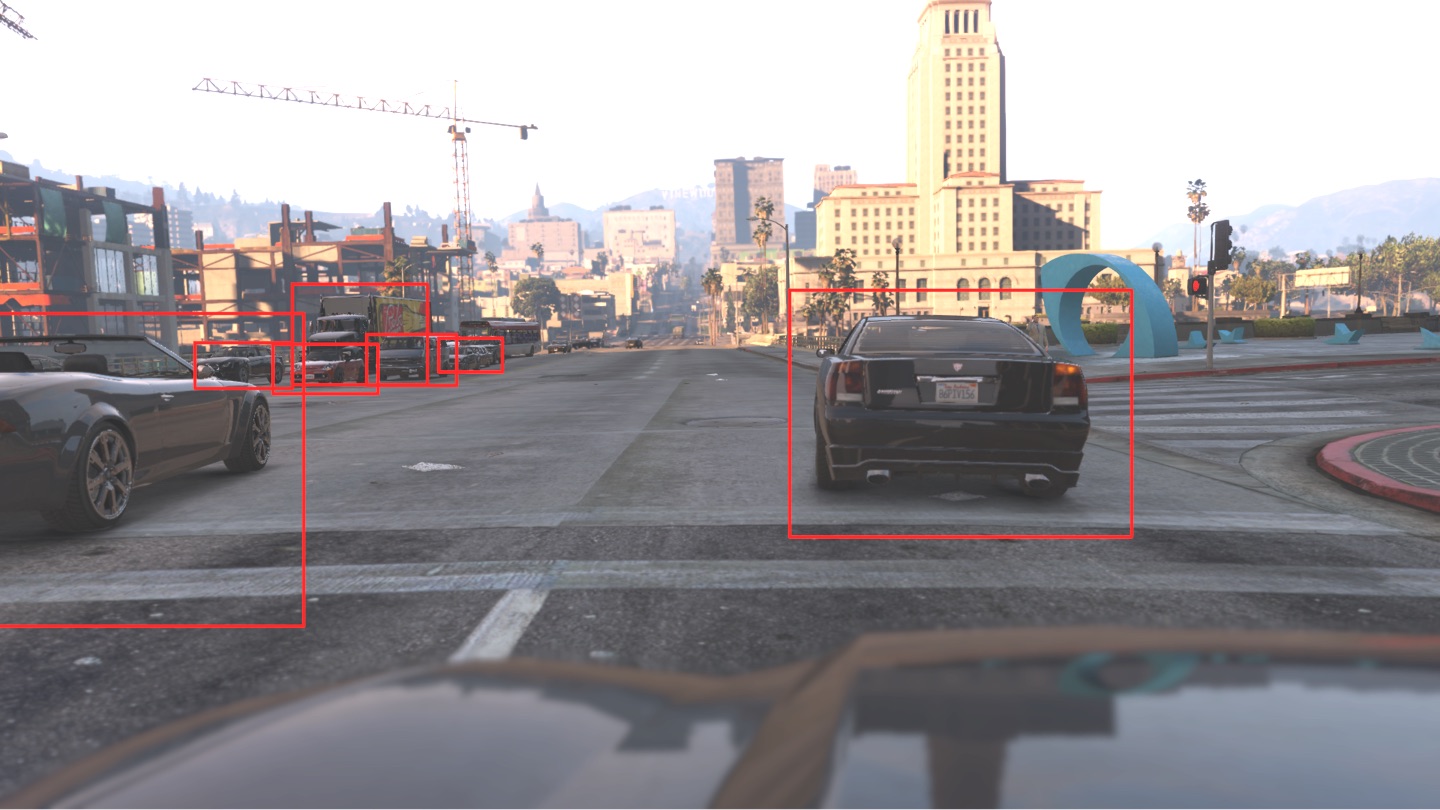}
	\includegraphics[width=0.49\textwidth]{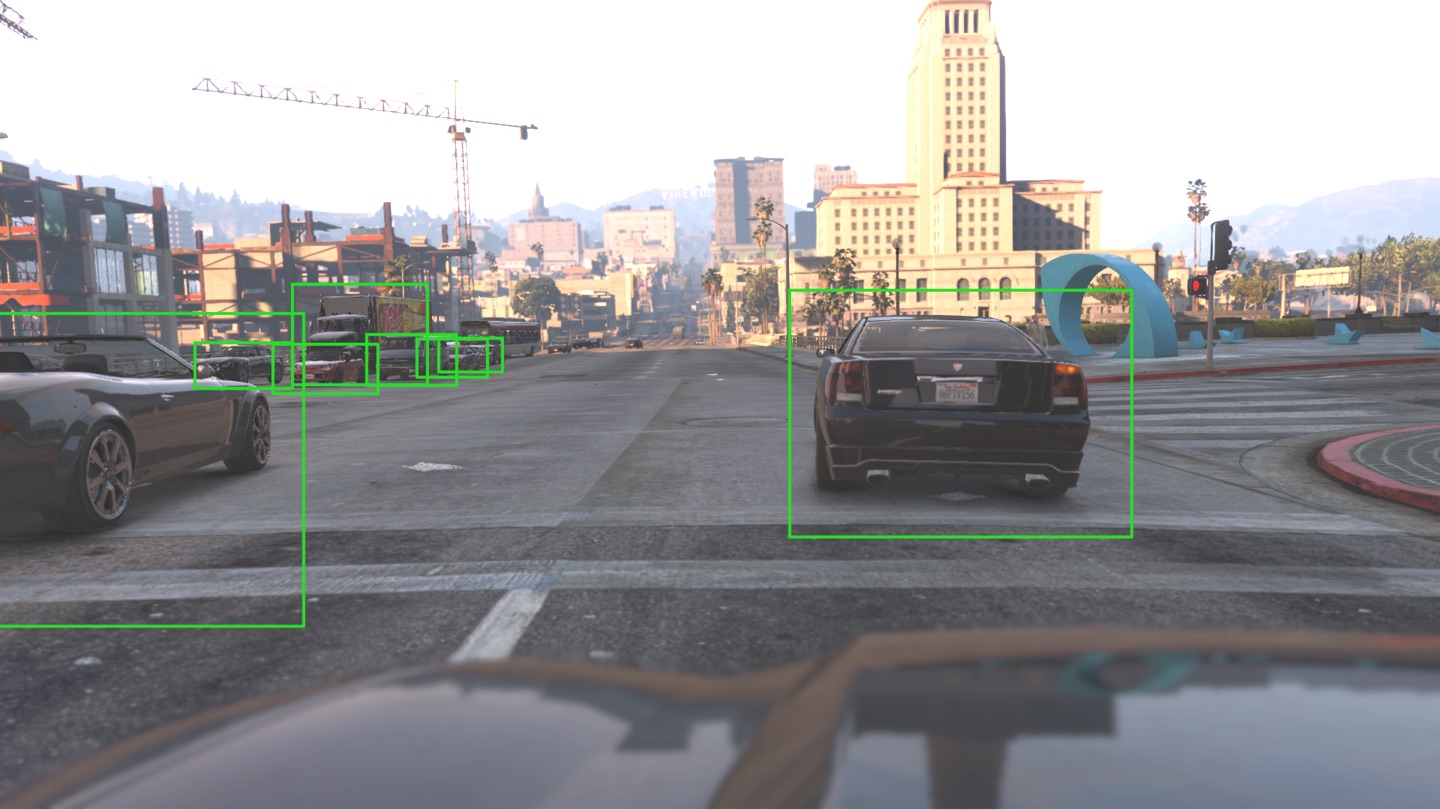}
	\includegraphics[width=0.49\textwidth]{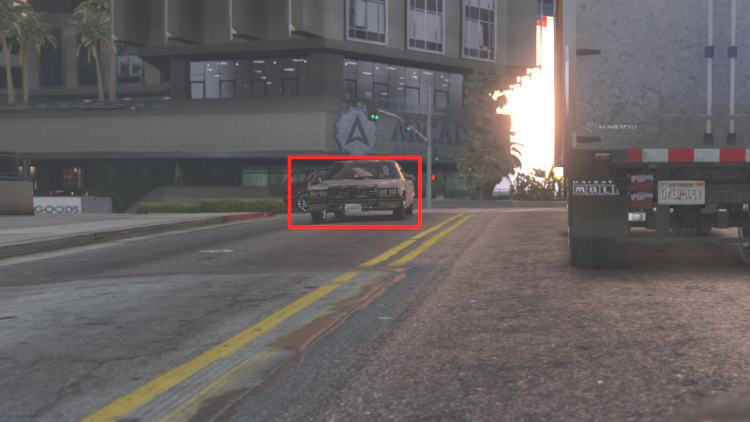}
	\includegraphics[width=0.49\textwidth]{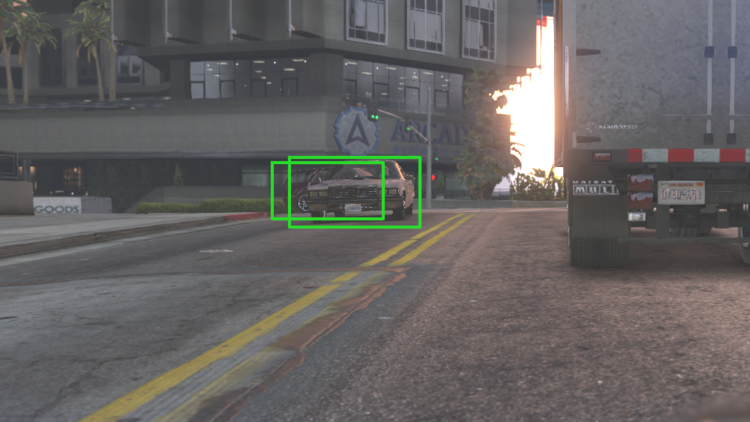}
	\caption{\textbf{Left:} Final detections without vg-NMS. \textbf{Right:} Final detections with vg-NMS. \label{fig:example_result}}
\end{figure}
\clearpage
\subsubsection{Synscapes}
\begin{figure}[h]
	\centering
	\includegraphics[width=0.49\textwidth]{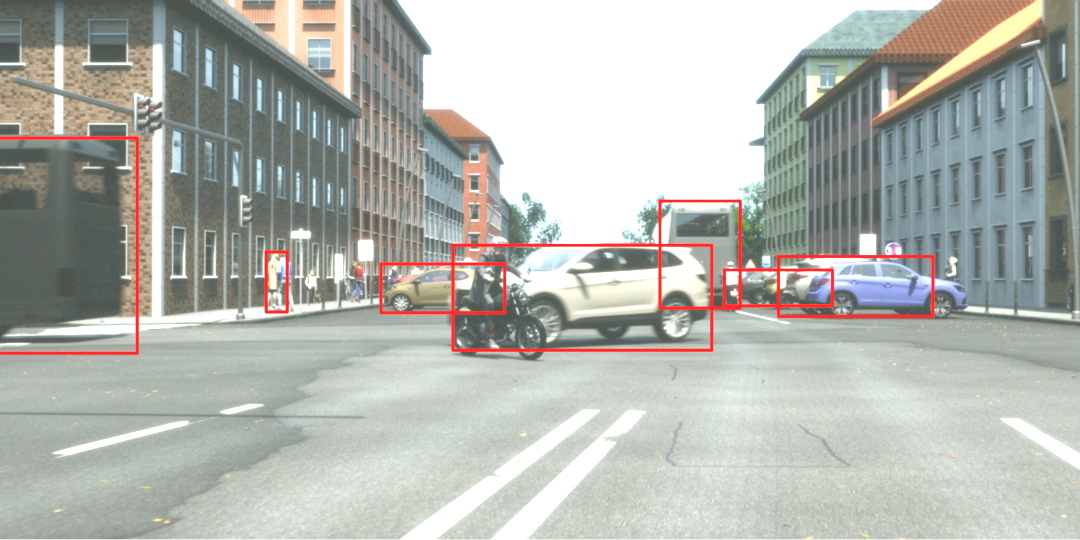}
	\includegraphics[width=0.49\textwidth]{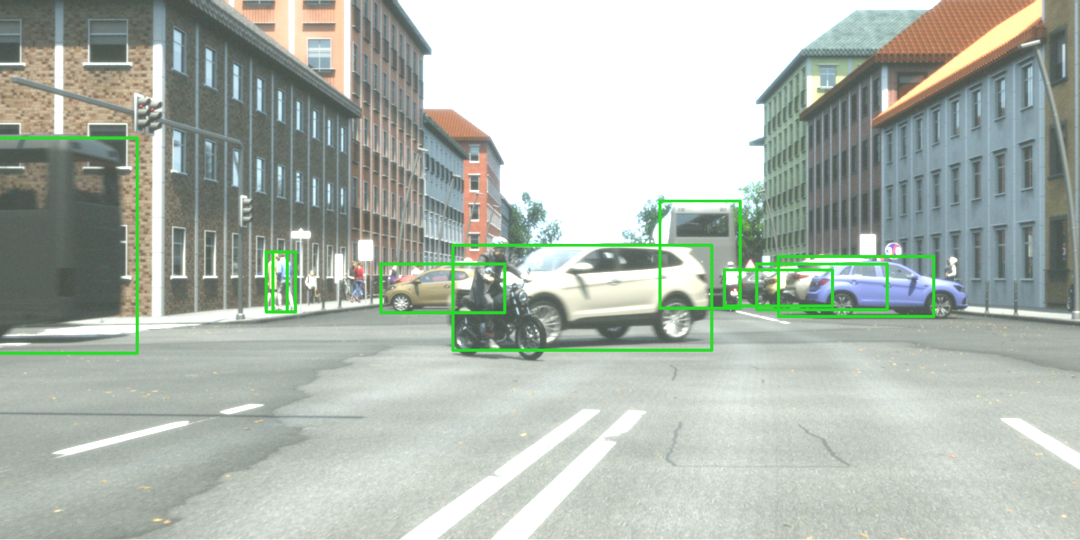}
	\includegraphics[width=0.49\textwidth]{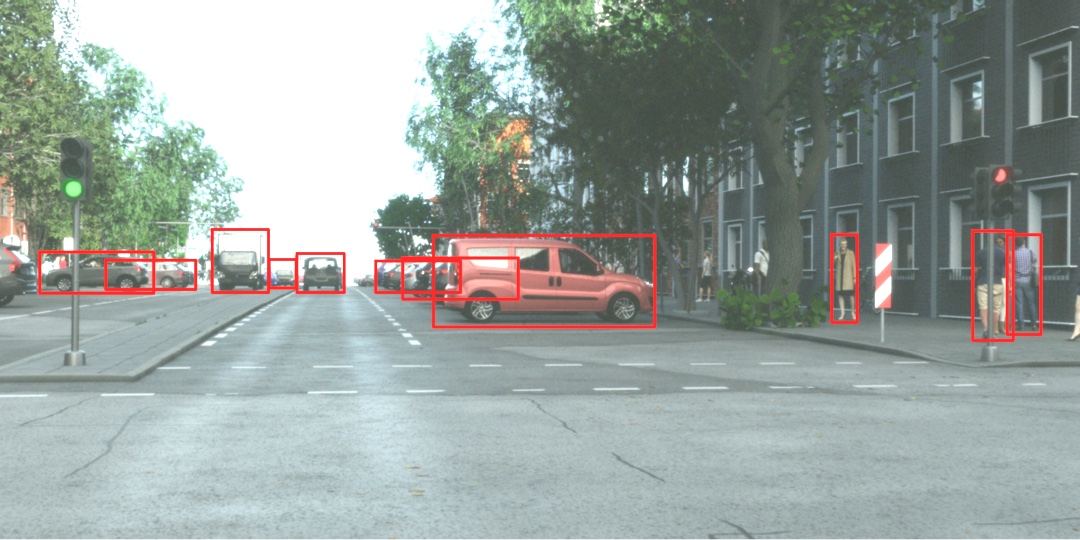}
	\includegraphics[width=0.49\textwidth]{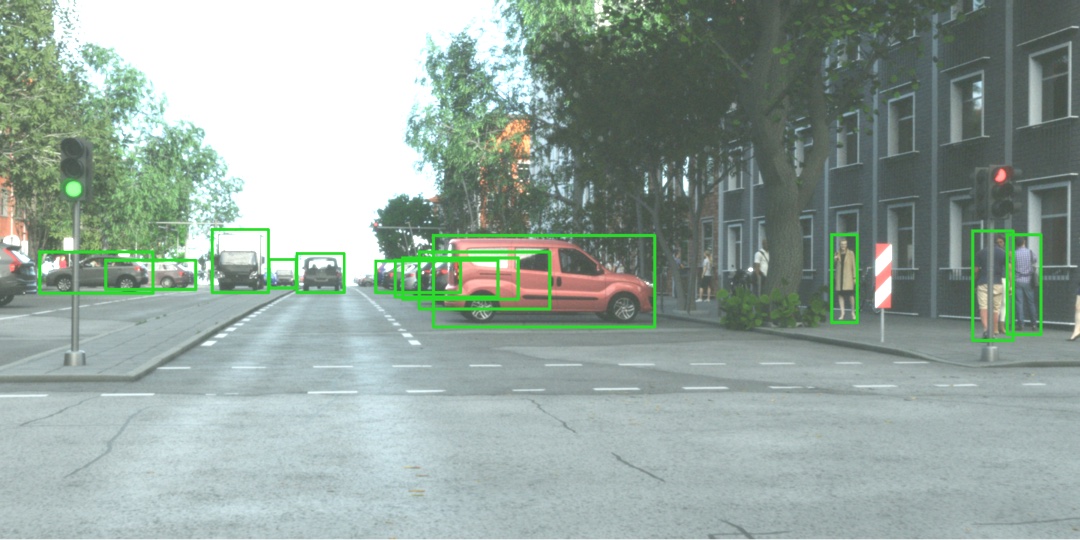}
	\caption{\textbf{Left:} Final detections without vg-NMS. \textbf{Right:} Final detections with vg-NMS. \label{fig:example_result}}
\end{figure}

\end{document}